\pdfoutput=1
\documentclass[11pt]{article}

\usepackage{acl}
\usepackage{graphicx}
\usepackage{multirow}
\usepackage{float}
\usepackage{amsmath}
\usepackage{times}
\usepackage{latexsym}
\usepackage{svg}

\usepackage[T1]{fontenc}
\usepackage[utf8]{inputenc}

\usepackage{microtype}

\usepackage{inconsolata}

\title{SKT5SciSumm - Revisiting Extractive-Generative Approach \\ for Multi-Document Scientific Summarization}

\author{
 \textbf{Huy Quoc To\textsuperscript{1,2}},
 \textbf{Ming Liu\textsuperscript{2}},
 \textbf{Guangyan Huang\textsuperscript{2}},
 \textbf{Hung-Nghiep Tran\textsuperscript{3}},\\
 \textbf{André Greiner-Petter\textsuperscript{4}},
 \textbf{Felix Beierle\textsuperscript{3,5}},
 \textbf{Akiko Aizawa\textsuperscript{3}}
\\
 \textsuperscript{1}University of Information Technology, VNU - HCM, Vietnam,\\
 \textsuperscript{2}Deakin University, Melbourne, VIC, Australia\\
 \textsuperscript{3}National Institute of Informatics, Tokyo, Japan,\\
 \textsuperscript{4}University of G\"{o}ttingen, G\"{o}ttingen, Germany,\\
 \textsuperscript{5}University of W\"urzburg, W\"urzburg, Germany
\\
  huytq@uit.edu.vn, \{q.to, m.liu, guangyan.huang\}@deakin.edu.au, \\\{nghiepth,aizawa\}@nii.ac.jp, greinerpetter@gipplab.org, felix.beierle@uni-wuerzburg.de
}

\begin{document}
\maketitle
\begin{abstract}
The summarization of scientific texts has shown significant benefits both for the research community and human society. Given the fact that the nature of scientific text is distinctive and the input of the multi-document summarization task is substantially long, the task requires sufficient embedding generation and text truncation without losing important information. To tackle these issues, in this paper, we propose SKT5SciSumm - a hybrid framework for multi-document scientific summarization (MDSS). We leverage the Sentence-Transformer version of Scientific Paper Embeddings using Citation-Informed Transformers (SPECTER) to encode and represent textual sentences, allowing for efficient extractive summarization using k-means clustering. We employ the T5 family of models to generate abstractive summaries using extracted sentences. SKT5SciSumm achieves state-of-the-art performance on the Multi-XScience dataset with 31.49\%, 8.23\%, 19.88\%, 33.23\% and 85.29\% for ROUGE-1,2,L,LSum and BERTScore, respectively. Our code is publicly shared on Github\footnote{The link will be available upon acceptance.}.
\end{abstract}

\section{Introduction}
The number of scientific documents has increased exponentially over the years. Although it is concrete proof that research activities are receiving more attention and emphasis, it creates a boundary for researchers to stay abreast of the latest advancements. The need for automatic summarization for scientific texts is inevitable. Although the single-document scientific summarization (SDSS) task requires the creation of an abstract for a paper using its content, the multi-document scientific summarization (MDSS) is proposed to conclude information from multiple topic-related papers \cite{ELKASSAS2021113679}.
\par Although pretrained language models have demonstrated impressive performance across various natural language processing tasks, there is still a lack of encoders specifically tailored for scientific text. As scientific text is usually written in a specific way and also contains academic phrases \cite{sugimoto-aizawa-2022-incorporating}, the encoder must be chosen appropriately to represent the text in the correct contextual embedding. Having good embeddings for scientific text is crucial to obtain decent performance in the text summarization task, as it requires the model to understand and summarize information \cite{beltagy-etal-2019-scibert}. Other problems include duplicate information, cross-document relationships, and longer text that MDSS models must deal with. 

\par To address these issues, we propose a hybrid method that embeds documents using SPECTER \cite{cohan2020specter}, extracts importance sentences using the k-mean algorithm, and summarizes the extracted sentences with a generative model - T5. In this approach, we present two phases of text summarization. An \textbf{unsupervised extractor} first narrows down those important sentences from the raw input text. This step helps eliminate irrelevant information and reduce the number of sentences. Then a \textbf{supervised abstractor} rewrites and further summarizes the output of the extractor. The abstractor is a generative model, in this work, we use T5 to produce the final summary that is close to the gold references. We fine-tune T5 with the extracted text from the train set and then evaluate it in the validation and test set.

\par To evaluate our proposed method, we use the Multi-XScience dataset \cite{lu-etal-2020-multi-xscience} which is the only large-scale and well-known dataset for MDSS. The task required in the dataset is to create a "related work" section by summarizing the abstract of a query paper and the abstracts of its referenced papers. Our empirical results show that: \textit{(1)} SKT5SciSumm achieves a noticeable improvement compared to other MDSS models in the Multi-XScience dataset, and \textit{(2)} T5-large version gives the best performance in the ROUGE score and BERTScore. Furthermore, on 50 random samples on test set, we query GPT-4 with zero-shot and few-shot prompting. As the results, our best model outperforms GPT-4's performance in both ROUGE scores and BERTScore.

\par In this work, we have two main findings: 
\begin{itemize}
  \item We propose a hybrid method - SKT5SciSumm which leverages both unsupervised extractive summarization using SPECTER encoder with K-means clustering and supervised abstractive summarization using T5 models for MDSS. Our proposed approach has proved to be simple yet efficient in multi-document scientific summarization tasks.
  \item This study compared the performance of various sizes of T5 models for MDSS. The results indicated that the combination of SPECTER, K-means clustering, and T5-large produced the highest ROUGE scores and BERTScore on the Multi-XScience dataset. T5-large is capable of capturing more intricate details and generates more logical and comprehensive summaries than its smaller counterparts. Although the T5-XL model has more parameters and is more advanced in other tasks, it was observed to paraphrase scientific phases and sentence structures in our experiments.
\end{itemize}

\section{Related Work}
In its early phases, MDSS research primarily concentrated on artificially generated small datasets \cite{hu-wan-2014-automatic, jaidka-etal-2013-deconstructing, hoang-kan-2010-towards}, employing unsupervised extractive techniques to extract sentences from multiple documents. The extractive summarization was made using purely statistical methods such as \cite{lexrank-erkan} or \cite{wan-yang-2006-improved}. \citet{mohammad-etal-2009-using} used citation information and summarization techniques to automatically generate a multi-document survey of scientific articles, to help researchers and scientists quickly understand large amounts of technical material. \citet{hoang-kan-2010-towards} introduced their prototype system, ReWoS, which uses a hierarchical set of keywords to drive the creation of an extractive summary. \citet{Jha_Coke_Radev_2015} proposed Surveyor - a system for generating coherent survey articles for scientific topics. The system uses an extractive summarization algorithm that combines a content model with a discourse model to produce coherent and readable summaries of scientific topics using text from a relevant scientific article. However, these unsupervised approaches face limitations in both capturing content and maintaining relationships, resulting in the challenge of generating high-quality summaries.

\par There are several attempts to make use of deep learning methods with large-scale datasets. \citet{wang-etal-2018-neural-related} presented a novel approach to automating the summarization of related work using a joint context-driven attention mechanism. The authors reported experimental results showing that this approach significantly outperforms a typical seq2seq summarizer and five classical summarization baselines. Another noticeable work was Relation-aware Related work Generator (RRG) proposed by \cite{chen-etal-2021-capturing}. Although this model used a Tranformer-based architecture for abstractive summarization, it was not able to create rich salient semantic summaries. Recently, \cite{shinde-etal-2022-extractive} proposed a method for multi-document summarization (MDS) of scientific documents that leverages both extractive and abstractive architectures. While this work demonstrates the merits of an extractive-then-abstractive approach for MDS, there are still some drawbacks that need to be addressed. For instance, their approach employs an outdated BERT-based extractive summarizer \cite{miller2019leveragingbertextractivetext}, which was trained on lecture notes. In contrast, we utilize the Sentence-BERT version of the SPECTER model, which was recently released and specifically trained on a large corpus of scientific texts, operating at the sentence level.

\par\citet{lu-etal-2020-multi-xscience} published the Multi-XScience dataset with several strong baselines that significantly contributed to the MDSS task. Since then, there has been some derivative research on this data set. PRIMERA \cite{xiao-etal-2022-primera} was designed to collect information in multiple documents, which is a crucial aspect in the summarization of multiple documents. However, in the Multi-XScience dataset, it underperformed the baselines. On the other hand, both REFLECT \cite{song-etal-2022-improving} and KGSum \cite{wang-etal-2022-multi} achieved competitive results using the extract-abstract framework. This proves that a hybrid approach containing both an extractor and an abstractor is appropriate for MDSS task. 

\section{Methodology}

\begin{figure*}[t]
\centering
    \includegraphics[scale=0.38]{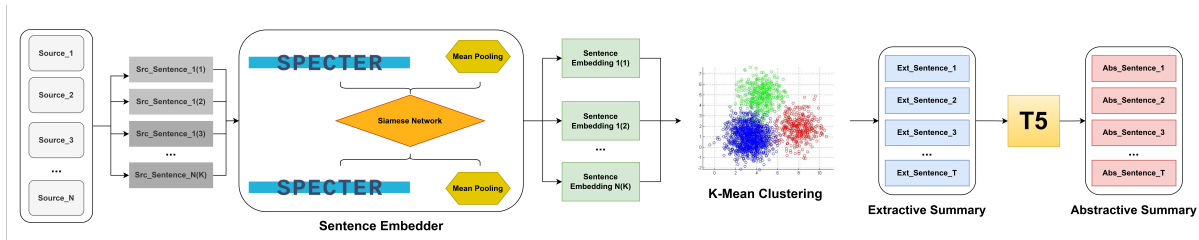}
    \caption{Our hybrid approach for multi-document scientific summarization.} 
    \label{fig:architecture}
\end{figure*}

SKT5SciSumm is created to generate comprehensive scientific summaries. It is able to identify key phrases and adhere to academic writing conventions. Our system is designed to address the task of writing a section of work using multiple sources. It combines all the documents, eliminates duplicates and irrelevant material, and produces concise summaries.
Our hybrid approach contains an \textbf{extractor} and an \textbf{abstractor}. The extractor consists of two components: SPECTER encoder and K-means clustering. We use the SPECTER sentence-transformer to create sentence embeddings and K-means clustering for choosing sentences to form an extractive summary. After that, we fine-tune the T5 model with extractive summaries. Figure \ref{fig:architecture} illustrates an overview of our approach.
\par In our extractor, we use the SPECTER \cite{cohan2020specter} model based on the Sentence-BERT architecture \cite{reimers2019sentencebert} that utilizes transformer-based deep learning techniques. It is pre-trained on a large corpus of scientific documents, allowing it to generate high-quality sentence embeddings that capture the semantic meaning and context of scientific sentences. These embeddings serve as dense vector representations of sentences, enabling efficient and effective processing of scientific text for various natural language processing tasks \cite{cohan2020specter}, including multi-document summarization. Meanwhile, K-means clustering \cite{Jin2010} is a popular unsupervised learning algorithm widely used to group data points into clusters based on their similarities. Combining these two methods enables us not only to represent the scientific sentences more accurately but also to choose the group sentences and then choose the most important one. 

\par On the other hand, the T5 model \cite{raffle-t5}, short for the "Text-to-Text Transfer Transformer," is a state-of-the-art language generative model. T5 is capable of performing various tasks, such as summarization \cite{rothe-etal-2021-thorough}, and question-answering \cite{lu-etal-2022-clinicalt5}, simply by converting the input into a textual format relevant to the specific task. With its encoder-decoder architecture, T5 has achieved impressive results in multiple benchmark datasets, demonstrating its versatility and effectiveness in various NLP tasks. In this paper, we also conduct a comprehensive study on T5 models in MDSS tasks by experimenting with four versions of T5, respectively, small (60M), base (220M), large (770M), and xl (3B)\footnote{Due to the GPU limitation, we are unable to fine-tune XXL (11B) version of T5 - which is also the largest one.}. 

\subsection{Extractor}

% \begin{figure*}[t]
% \centering
%     \includegraphics[scale=0.3]{paclic2023/extractor.png}
%     \caption{The silhouette score is a metric used to evaluate the quality of clustering results in k-means clustering. Measures how well each data point fits within its assigned cluster compared to other clusters.} 
%     \label{fig:extractor}
% \end{figure*}

Our strategy for an extractor is to generate the embeddings of documents in a group using SPECTER \cite{cohan2020specter}, then use K-means to choose the most important sentences. What are important sentences in the context of MDS? These sentences should contain rich and condensed information that covers most of the context. An overview of our extractor is shown in Figure \ref{fig:architecture}. Although clustering-based methods have been studied since the early 2000s \cite{RADEV2004919, wang2008-multisum}, they still prove to be an efficient method for the multi-document summarization task. In \citet{ernst-etal-2022-proposition} work, the authors suggest a method that involves taking out propositions from the input documents, discarding non-important propositions, categorizing salient propositions based on their semantic similarity, and combining the clusters to create summary sentences. Meanwhile, our extractor focuses on document-level embeddings using SPECTER and academic structures with the documents. Therefore, our approach is capable of extracting scientific structures and choosing salient academic sentences.

  \subsubsection{SPECTER Embeddings}
 SPECTER \cite{cohan2020specter} is a new method to generate document-level embeddings of scientific documents based on pretraining a Transformer language model on the citation graph. Additionally, SPECTER is applicable in situations where metadata, such as authors or venues, are not available. SPECTER uses citations as a naturally occurring, inter-document incidental supervision signal indicating which documents are most related, and formulates the signal into a triplet-loss pre-training objective. This allows SPECTER to incorporate inter-document context into the language model and learn document representations. It is designed to be easily applied to downstream applications without task-specific fine-tuning and has shown substantial improvements over a wide variety of baselines. Therefore, it is suitable for our unsupervised approach as an extractor. In our experiments, we employ the sentence-transformer version of SPECTER, as we aim to encode each sentence in the document. 

 \subsubsection{Clustering embedding}
K-means clustering \cite{Jin2010} is a simple and efficient unsupervised algorithm that is capable of handling large amounts of text data. It automatically groups similar sentences together, allowing the extraction of the most representative sentences from a document cluster as summarization candidates. The scalability of the algorithm makes it suitable for real-time and large-scale summarization. To obtain the extractive summary, we choose the sentences in centered positions (centroids) of each cluster. The drawback of this method is that it requires a redefined number of \textbf{\textit{K}} which may cause suboptimal results as the number of input sentences is different. To address this problem, we first calculate the silhouette score to obtain the optimal \textbf{\textit{K}} for each input string. Silhouette scoring offers a comprehensive evaluation of cluster quality considering both cohesion and separation of data points within and between clusters, respectively. Higher silhouette scores indicate well-defined and distinct clusters, whereas lower scores suggest that data points might fit better in other clusters. By computing the silhouette score for various values of \textbf{\textit{K}}, we can identify the value that produces the highest score, thus identifying the ideal number of clusters for the dataset. Since we want the summary to have at least two sentences from \(T\) input sentences in one document, the range of \textbf{\textit{K}} is: 
\[
{K} = [2, \frac{T}{2} ]
\]
This ensures that our model can handle both extremely short and long input text. The final step is to concatenate all summaries of each document to form the final extractive summary for a set of documents \(D\). Having the extractor in a multi-document summarization is an advancement that helps reduce duplicate information and choosing keywords for the abstractor.

\subsection{Abstractor}

We chose T5 as our abstractor for a number of reasons. First, this research aims to study how generative language models perform in summarizing scientific articles. T5 \cite{raffle-t5} and BART \cite{lewis-etal-2020-bart} are two well-known generative models that have shown their efficiency in generating summaries in many general and other specific domains. Regarding scientific domains, BART has been studied and achieved noticeable results. Therefore, we put T5 under experiments not only to explore its performance compared to BART, but also to examine whether text-to-text architecture is capable of generating decent summaries in the scientific domain. 
To implement the T5 model, we first fine-tune the model with train and validation sets. The datasets used for fine-tuning are extractive summaries retrieved from the extractor. As mentioned above, the extractive summaries contain only important sentences that are more effective for fine-tuning to generate more condensed summaries. 

\section{Experiments}
 In this section, we first analyze the Multi-XScience dataset to gain more insights. Then we briefly describe the ROUGE and BERTScore metrics that are used to evaluate the results. Finally, we present our experimental setting in detail. 
\begin{figure*}[t]
\centering
    \includegraphics[scale=0.19]{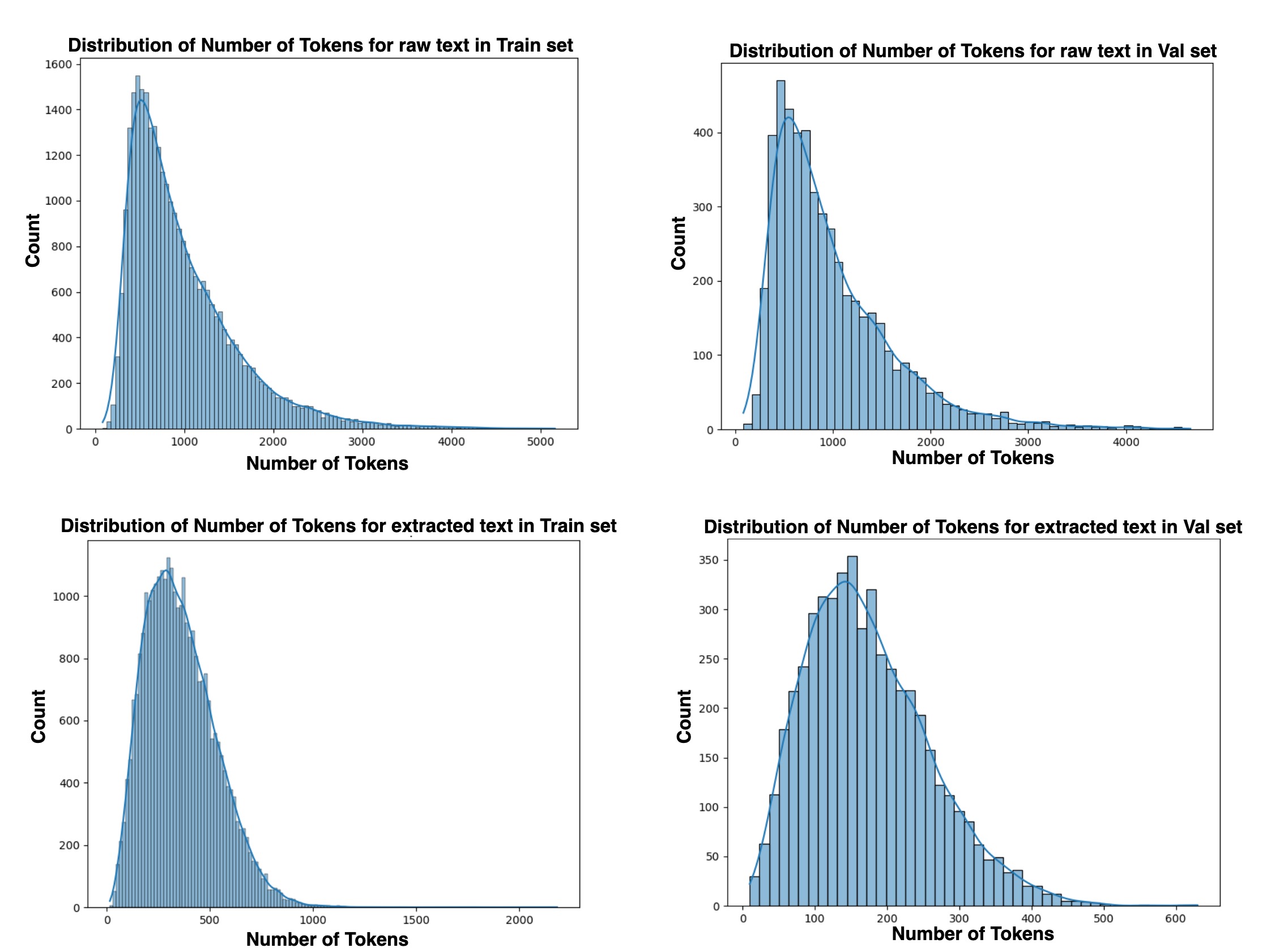}
    \caption{Distribution of tokens from raw input text  compared with extracted summaries in train and validation set using T5 tokenizer.} 
    \label{fig:distribution_raw}
\end{figure*}

\subsection{Dataset}
 The Multi-Xscience dataset \footnote{\url{https://github.com/yaolu/Multi-XScience}} \cite{lu-etal-2020-multi-xscience} is an open-source large-scale multi-document summarization dataset created from scientific articles in English. It introduces a challenging multi-document summarization task: writing the related work section of a paper based on its abstract and the articles it references. The dataset was created using a dataset construction protocol called extreme summarization, which favors abstractive modeling approaches. Additionally, Multi-XScience contains fewer positional and extractive biases than previous multi-document summarization datasets, making it more challenging and requiring models with a high level of text abstractiveness. The Multi-XScience dataset contains a total of 40,528 documents, divided into three sets: 30,369 for training, 5,066 for validation, and 5,093 for testing.
 \par Several models were used to test the effectiveness of Multi-XScience, including two commonly used unsupervised extractive summarization models, LexRank \cite{lexrank-erkan} and TextRank \cite{mihalcea-tarau-2004-textrank}, as baselines. For supervised abstractive models, HiMAP \cite{fabbri-etal-2019-multi} and HierSumm \cite{liu-lapata-2019-hierarchical} were tested. Both models deal with multi-documents using a fusion mechanism, which performs the transformation of the documents in the vector space. HiMAP adapts a pointer-generator, while HierSumm uses a hierarchical encoder-decoder architecture. BART \cite{lewis-etal-2020-bart} was also evaluated as a baseline model that achieved competitive results.
 \par We further analyze the dataset using the T5 tokenizer to see the length of one input. Figure \ref{fig:distribution_raw} shows the distribution of input length in train and validation sets. It clearly illustrates that most of the inputs have around 1000 tokens. However, there are still some cases in which the length can be up to 4000 tokens, which can be challenging for T5 to handle. 

\subsection{Evaluation Metrics}
In this article, we report the ROUGE \textit{F1 score} \cite{lin-2004-rouge} to evaluate the performance of our proposed method. Although ROUGE has been widely used to access summarization models, there remain some ambiguous points for ROUGE-L among previous research, especially on the Multi-XScience dataset. 
\par As stated in \cite{lin-2004-rouge}, ROUGE-L is an automatic summarization evaluation method that measures the Longest Common Subsequence (LCS) between a candidate summary and a set of reference summaries. It takes the union LCS score, which means that it considers all the common subsequences between the candidate and reference summaries, rather than just the longest one. There are two approaches to calculate ROUGE-L which are sentence-level LCS and summary-level LCS. Sentence-level LCS computes the LCS between two summary sentences, while summary-level LCS computes the LCS between a reference summary and a candidate summary. To compute the summary-level LCS, the union LCS matches between a reference summary sentence and every candidate summary sentence are taken. In our experiments, we compute the ROUGE-L score on both the sentence-level and summary-level. 

\par Additionally, in our evaluation, we also run BERTScore \cite{Zhang2020BERTScore} to measure the similarity between the generated text and reference text. While ROUGE metrics only calculate the similarity of two given texts by considering their n-gram overlaps, BERTScore is measured based on the cosine similarity between two pieces of texts using their contextual embeddings. 
\subsection{Implementation Details}
As discussed in the previous section, we first concatenate all source texts from one document into one paragraph. Then it runs through an extractor to deduce all irrelevant information. This step also decreases the number of sentences, leading to improved fine-tuning results. Figure \ref{fig:distribution_raw} shows the length of the input text after being processed by our extractor. Most of the extracted summaries have less than 1000 tokens, which is ideal for training the T5 model.
\begin{table}[H]
\footnotesize
\centering
\resizebox{\columnwidth}{!}{%
\begin{tabular}{l|ccccc}
\hline
\textbf{Model} &  \textbf{Size} & \textbf{Lr} & \textbf{Batch size} & \textbf{GrA}  \\ \hline
\multirow{4}{*}{T5} &  small & 1e-5  & 32 & 1 \\ 
                    &  base & 1e-5 & 8 & 4 \\ 
                    &  large & 1e-6 & 4 & 8 \\ 
                    &  xl & 1e-7 & 1 & 32 \\ \hline
\end{tabular}%
}
\caption{Experimental settings for T5 models.}
\label{tab:t5_settings}
\end{table}
% \begin{figure*}[t]
% \centering
%     \includegraphics[scale=0.42]{distribution_ext.png}
%     \caption{Distribution of extracted summaries in train and validation set using T5 tokenizer} 
%     \label{fig:distribution_ext}
% \end{figure*}

\par We set up our experiment to run on a single NVIDIA A100 GPU with 80GB of VRAM. Due to this limitation, we can only fine-tune four versions of T5: small (60M), base (220M), large (770M), and xl (3B). These versions are available at huggingface\footnote{\url{https://huggingface.co/docs/transformers/model_doc/t5}}. We describe our training settings in Table \ref{tab:t5_settings}. Since our experimental target is to have similar configurations for all models, we adapt the gradient accumulation (GrA) to simulate the same batch size of 32 and train for 8 epochs. However, the learning rates (Lr) still have to be adjusted accordingly to avoid over-fitting. We set the weight decay to 0.2 and save the top-3 checkpoints based on the evaluation results on the validation set. Based on the number of tokens in the reference text using T5, we set the output length for the T5 models to 256 tokens to match the desired reference length. The remaining parameters are left as default settings. 

\par We finally evaluate the fine-tuned T5 models in the test set. For each version of T5 and each test sample, we generate 5 different summaries. The objective is to investigate the consistency of the generative models. We then measure the generated results on the average ROUGE F1-score and average F1-BERTScore for comparison. 

\section{Results}
Table \ref{tab:t5_comparison} summarizes the results of our approach on the test set on four types of ROUGE scores. It clearly indicates that the large version of T5 with 770M parameters achieves the best performance compared to other versions. Noticeably, the T5-xl version, though it has almost four times more parameters than the T5-large model, its results are slightly lower than those of T5-large. Specifically, in ROUGE-1, ROUGE-L, and ROUGE-LSum scores, T5-large is, respectively, 0.17\%, 0.08\% and 0.11\% better than T5-xl.
\begin{table}[H]
\resizebox{\columnwidth}{!}{%
\centering
\begin{tabular}{l|cccccc}
\hline
\textbf{Model}               &  \textbf{Size} & \textbf{R-1} & \textbf{R-2} & \textbf{R-L} & \textbf{R-LSum} & \textbf{BERTScore}  \\ \hline
\multirow{4}{*}{T5} &  small & 36.92 & 7.90 & 19.46 & 32.18 & 85.11 \\ 
                    &  base & 37.20 & 8.23 & 19.76 & 32.53 & 85.28 \\ 
                    &  \textbf{large} & \textbf{37.49} & \textbf{8.65} & \textbf{19.88} & \textbf{33.23} & \textbf{85.29} \\ 
                    &  xl & 37.32 & 8.65 & 19.80 & 33.02 & 85.29 \\ \hline
\end{tabular}%
}
\caption{Comparison of different size of T5 models. \textbf{R} is the abbreviation of ROUGE.}
\label{tab:t5_comparison}
\end{table}

In the BERTScore evaluation, all four models achieved fairly similar scores. Although T5-large obtains the highest score, the difference gap is only around 0.1\%. 

\begin{table*}[t]
\centering

\begin{tabular}{l|cccc}
\hline
\textbf{Model} & \textbf{Rouge-1} & \textbf{Rouge-2} & \textbf{Rouge-L} & \textbf{Rouge-LSum} \\ \hline
Hiersumm\textsuperscript{*} & 30.02 & 5.04 & - & 27.60 \\
HiMAP\textsuperscript{*} & 31.66 & 5.91 & - & 28.43 \\
BertABS\textsuperscript{*} & 31.56 & 5.02 & - & 28.05 \\
BART\textsuperscript{*} & 32.83 & 6.36 & - & 26.61 \\
SciBertABS\textsuperscript{*} & 32.12 & 5.59 & - & 29.01 \\ 
Pointer-Generator\textsuperscript{*} & 34.11 & 6.76 & - & 30.63 \\ \hline
PRIMERA \cite{xiao-etal-2022-primera} & 31.93 & 7.37 & 18.02 & - \\ \hline
REFLECT \cite{song-etal-2022-improving} & 34.18 & 8.20 & 17.42 & 29.73 \\ \hline
KGSum \cite{wang-etal-2022-multi} & 35.77 & 7.51 & - & 31.43 \\ \hline
\textbf{SKT5SciSumm (Ours)} & \textbf{37.49} & \textbf{8.23} & \textbf{19.88} & \textbf{33.23} \\ \hline
\end{tabular}
\caption{Performance of our approach compared to baselines and related works. The results with \textsuperscript{*} are retrieved from \cite{lu-etal-2020-multi-xscience}.}
\label{tab:comparision_results}
\end{table*}

\par In addition, given that the Multi-XScience is a large-scale dataset and the T5-770M and T5-3B models are fairly large language models, it takes more time for training and inference, yet the performance is not too far from the smallest version of T5. For example, the margin that T5-large achieves on the ROUGE-1 score is only 0.57\% higher compared to T5-small, while its training phase is four times longer.

\par We compare our best results with baselines and other previous abstractive summarization models in Table \ref{tab:comparision_results}. Our proposed method outperforms previous models in all ROUGE metrics on the Multi-XScience dataset. Compared to the predecessor state-of-the-art model, KGSum, our best model achieves remarkably higher scores. Specifically, our improvements are 1.72\%, 0.74\% and 1.8\% in ROUGE-1, ROUGE-2, and ROUGE-LSum, respectively. Based on the given code, while PRIMERA \cite{xiao-etal-2022-primera} used ROUGE-L (sentence-level LCS)\footnote{\url{https://github.com/allenai/PRIMER}}, KGSum \cite{wang-etal-2022-multi} evaluated ROUGE-LSum (summary-level LCS)\footnote{\url{https://github.com/muguruzawang/KGSum}}. To our best knowledge, the evaluation code for all baseline models from \cite{lu-etal-2020-multi-xscience} is not available. Therefore, in Table 3, we follow \cite{wang-etal-2022-multi} and consider the ROUGE-L score from the \citet{lu-etal-2020-multi-xscience} baselines as ROUGE-LSum (summary-level). In addition, we are not able to compare our BERTScore with other models since it was not measured in the previous works. 

\section{Discussion}
% To further prove the effectiveness of SKT5SciSumm, we conduct some ablation experiments for our extractor and abstractor on the Multi-XScience dataset. 
\subsection{Ablation study}
The goal of our ablation study is to assess the performance of SPECTER with K-means (SK) clustering individually. We evaluate our extractor in the test set and compare it with other extractive summarization methods and report in Table \ref{tab:ablation_ext}. Our extractor gives better results compared to the former extractive approaches. In ROUGE-1, ROUGE-2, and ROUGE-LSum, respectively, we improve at least by 2.10\%, 1.46\%, 1.57\%. SK-extractor scores are also competitive compared to Ext-Oracle\footnote{\url{https://pypi.org/project/extoracle/}}, which creates extractive upper bound results.
\begin{table}[H]
\centering
\resizebox{\columnwidth}{!}{%
\begin{tabular}{l|ccc}
\hline
\textbf{Model} & \textbf{R-1} & \textbf{R-2} & \textbf{R-LSum} \\ \hline
LEAD\textsuperscript{*} & 27.46 & 4.57 & 18.82  \\
LexRank\textsuperscript{*} & 30.19 & 5.53 & 26.19  \\
TextRank\textsuperscript{*} & 31.51 & 5.83 & 26.58 \\
\textbf{SPECTER+K-means (SK)}& \textbf{33.61} & \textbf{7.29} & \textbf{28.15}\\ \hline
Ext-Oracle\textsuperscript{*} & 38.45 & 9.93 & 27.11\\   \hline                       
\end{tabular}%
}
\caption{Performance of our extractor compared to other extractive methods. The results with \textsuperscript{*} are retrieved from \cite{lu-etal-2020-multi-xscience}. ROUGE-L was not available in their work.}
\label{tab:ablation_ext}
\end{table}
\subsection{Comparison with GPT-4} \label{sec:compare_w_gpt4}
To further investigate our proposed method with one of the state-of-the-art large language models, we use OpenAI API\footnote{\url{https://platform.openai.com/docs/models/gpt-4-turbo-and-gpt-4}} to query GPT-4 in two settings: zero-shot prompting, and few-shot prompting. In each setting, we also evaluate GPT-4 further by passing to the query full text and extracted text from our extractor respectively. Particularly, in zero-shot prompts, we directly pass the source text and ask the model to generate the summaries. For few-shot prompting, we give GPT-4 with 1-3 example pairs and then query for answers. Due to cost restrictions, we only examine 50 random samples from the test set and compare GPT-4's performance with our method.
\begin{table*}[t]
\centering
\begin{tabular}{l|ccccc}
\hline
\textbf{Model} & \textbf{R-1} & \textbf{R-2} & \textbf{R-L} & \textbf{R-LSum} & \textbf{BERTScore} \\ \hline
GPT-4 zs-ft & 28.73 & 4.41 & 14.16 & 25.10 & 82.98 \\
GPT-4 fs-ft & 28.85 & 4.59 & 13.70 & 25.31 & 82.99 \\
GPT-4 zs-ext & 29.67 & 4.61 & 14.89 & 26.13 & 83.91 \\
GPT-4 fs-ext & 30.58 & 5.04 & 15.00 & 26.81 & 84.06\\
\textbf{SKT5SciSumm}& \textbf{36.65} & \textbf{6.57} &\textbf{18.75} & \textbf{31.90} &\textbf{84.83}  \\ \hline                    
\end{tabular}
\caption{Performance of our method compared to GPT-4 on 50 random samples from test set. zs is short for zero-shot, fs is short for few-shot, ft is short for full text and ext is short for extracted text.}
\label{tab:gpt}
\end{table*}
\par The results in Table \ref{tab:gpt} indicate that our SKT5SciSumm method clearly outperforms GPT-4. Our approach surpasses GPT-4, respectively, by around 6\%, 2\%, 4\%, and 5\% on ROUGE1,2, ROUGE-L and ROUGE-LSum score. However, BERTScore results of GPT-4 are only 0.6\% lower than our best results. This implies that GPT-4 rewrites the input text and replaces it with synonyms or related words. Based on the above observation, even though the summaries generated by GPT-4 have similar overall meaning compared to ours, they have different vocabulary and phrasing compared to the reference summaries. One possible explanation is the fact that our models are well fine-tuned on scientific text, whereas GPT-4 is predominantly trained on a broader range of domains. Additionally, the improved performance of GPT-4 using extracted text confirms the effectiveness of our extractor in generating more concise information for summarization. For instance, in the few-shot setting, by using the extracted text, the performance of GPT-4 is increased by 1.73\%, 0.45\%, 1.30\%, 1.50\%, and 1.07\% on ROUGE-1, 2, L, LSum and BERTScore respectively.

\subsection{Factual consistency evaluation}
To further validate our results, we perform a factual consistency check using AlignScore \cite{zha-etal-2023-alignscore}. This metric compares the generated summaries and the original text to examine whether generative models create hallucinations when summarizing the documents.

\begin{table}[H]
    \centering
    \begin{tabular}{l|c} \hline
      \textbf{T5 Model}  & \textbf{AlignScore}\\ \hline
      small   &  85.14\\
      base   & 86.25\\
      \textbf{large} & \textbf{90.36}\\
      xl   & 90.33\\ \hline
    \end{tabular}
    \caption{Factual consistency evaluation on four versions of fine-tuned T5}
    \label{tab:alignscore}
\end{table}

The results in Table \ref{tab:alignscore} demonstrate that the summaries generated from our models have a minimal percentage of hallucinations and remain highly consistent with the original input documents. 
\subsection{Result Analysis}
We perform a human evaluation on the summaries generated by our method and GPT-4. The detail of the human analysis is in the Appendix \ref{sec:appendix_b}. In addition, we review some examples generated by four versions of T5 to investigate how the summaries differ from each other. Table \ref{tab:results_analysis} in the Appendix \ref{sec:appendix_a} shows one instance of the test set. In the table, we find that our fine-tuned T5 models are able to capture the correct keywords. However, the large and xl versions of T5 generate more coherent summaries while maintaining the salient information. Therefore, their results are similar to those of human writing. In this analysis, we also notice that T5-xl captures a good number of academic phrases. However, compared to the T5-large version, most of the academic structures have been rewritten. Hence, its performance on the ROUGE score is slightly lower.

\section{Conclusion}
In this paper, we present SKT5SciSumm, a hybrid generative method for MDSS. Our model utilizes the power of SPECTER and K-means clustering to handle long and complicated documents, and generates proficient summaries. Experimental results show that our proposed model outperforms all baselines and previous multi-document summarization methods; hence, it achieves state-of-the-art results on the Multi-XScience dataset. Our approach yields the fact that, by leveraging simple and well-known techniques, it is able to produce better results compared to the previous complicated systems on the MDSS task. The efficiency of our method is also demonstrated by comparing its results with GPT-4 under automatic and human evaluation. Future work is possible, but not limited, to further explore the performance of other generative models in the processing of scientific text. We are also curious to explore the performance of mT5 for the MDSS task in other languages.

\section{Limitations}
Due to GPU limitations, we are unable to evaluate the largest version of T5 (XXL - 11B). Since our scope is to propose a method for multi-document summarization on scientific text, SKT5SciSumms is not evaluated on other open-domain datasets. With that being said, the combination of extractive and abstractive methods is applicable for most of the multi-document summarization. Moreover, considering that the proposed framework is fine-tuned and GPT-4 is not fine-tuned, the comparison proposed in section \ref{sec:compare_w_gpt4} and Appendix \ref{sec:appendix_b} has some minor drawbacks. For example, the fine-tuning process applied to the proposed framework likely optimizes it for specific tasks or datasets, making it more tailored to those contexts. In contrast, GPT-4, being a general model without such fine-tuning, might not perform as well on these specific tasks, potentially skewing the comparison.
% Remove for double blind
% \section*{Acknowledgements}
% This work was partially funded by the German Academic Exchange Service (DAAD) - 57515245.


\bibliography{anthology,refs}
\appendix

\section{Human Evaluation}
\label{sec:appendix_b}
To validate the results generated by SKT5SciSumm and GPT-4, we evaluated 50 samples that were randomly selected for GPT-4 in Section \ref{sec:compare_w_gpt4}. We created a questionnaire for two Ph.D. students, asking them to: (i) choose which summary is the most similar to the reference, and (ii) score both summaries in terms of relevance and readability on a scale from 1 to 5. The relevance and readability of the generated text were determined by asking the evaluators two questions:
\begin{itemize}
    \item To what extent do you think this text is relevant to the given reference text?
    \item To what extent do you think this text is fluent compared to the given reference text?
\end{itemize}

\begin{figure}[h]
\centering
    \includegraphics[scale=0.35]{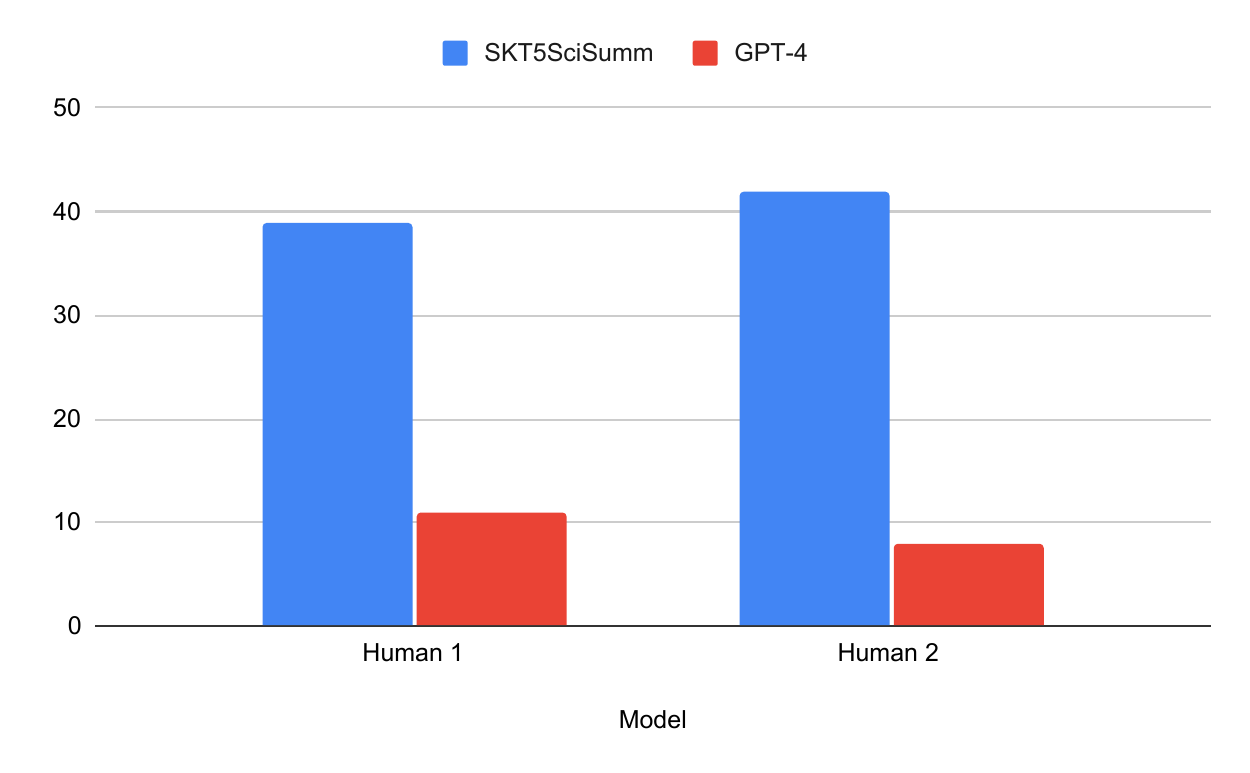}
    \caption{The voting results of two humans on generated resutls of SKT5SciSumm and GPT-4 compared to references.} 
    \label{fig:votes}
\end{figure}

In the first task, we summarize the votes of the two students in Figure \ref{fig:votes}. The figure clearly demonstrates that both evaluators believed that the summaries generated by our method are more similar to the provided references.

\begin{figure}[h]
\centering
    \includegraphics[scale=0.35]{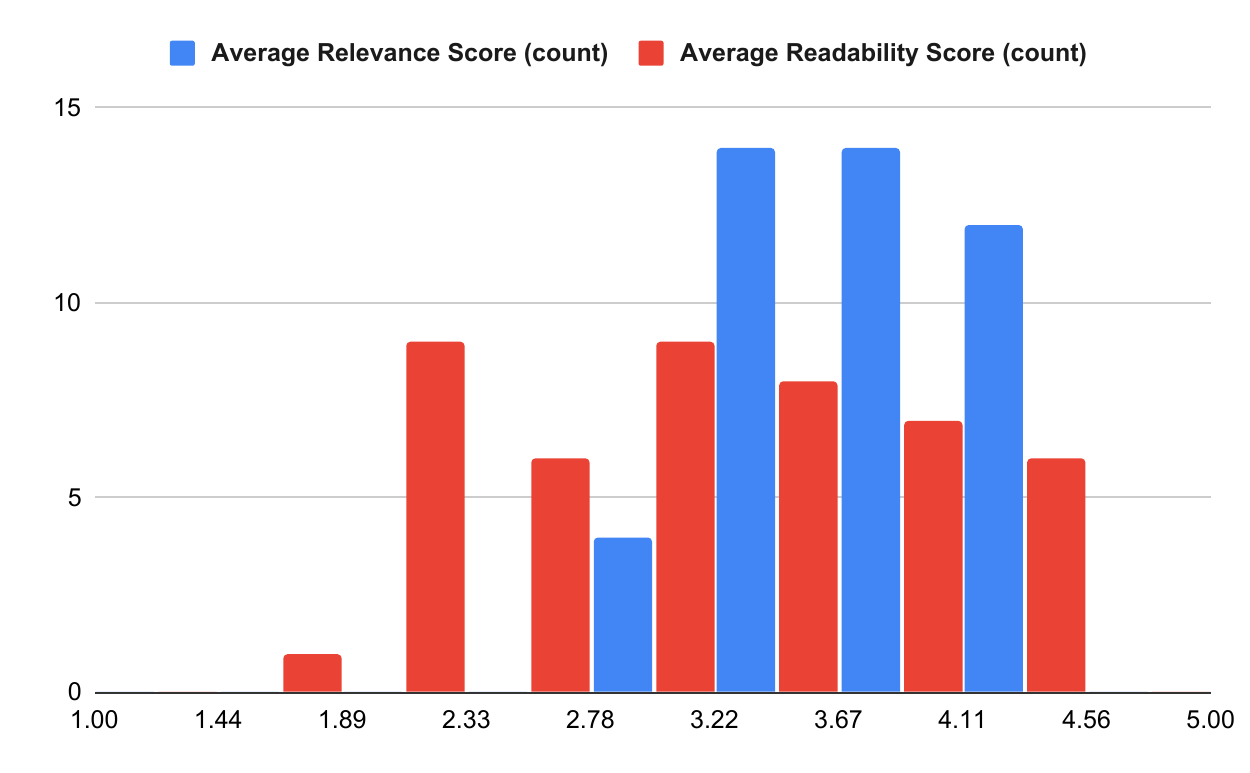}
    \caption{The distribution of average relevance and readability scores for summaries generated by SKT5SciSum.} 
    \label{fig:scores-skt5}
\end{figure}

\begin{figure}[h]
\centering
    \includegraphics[scale=0.35]{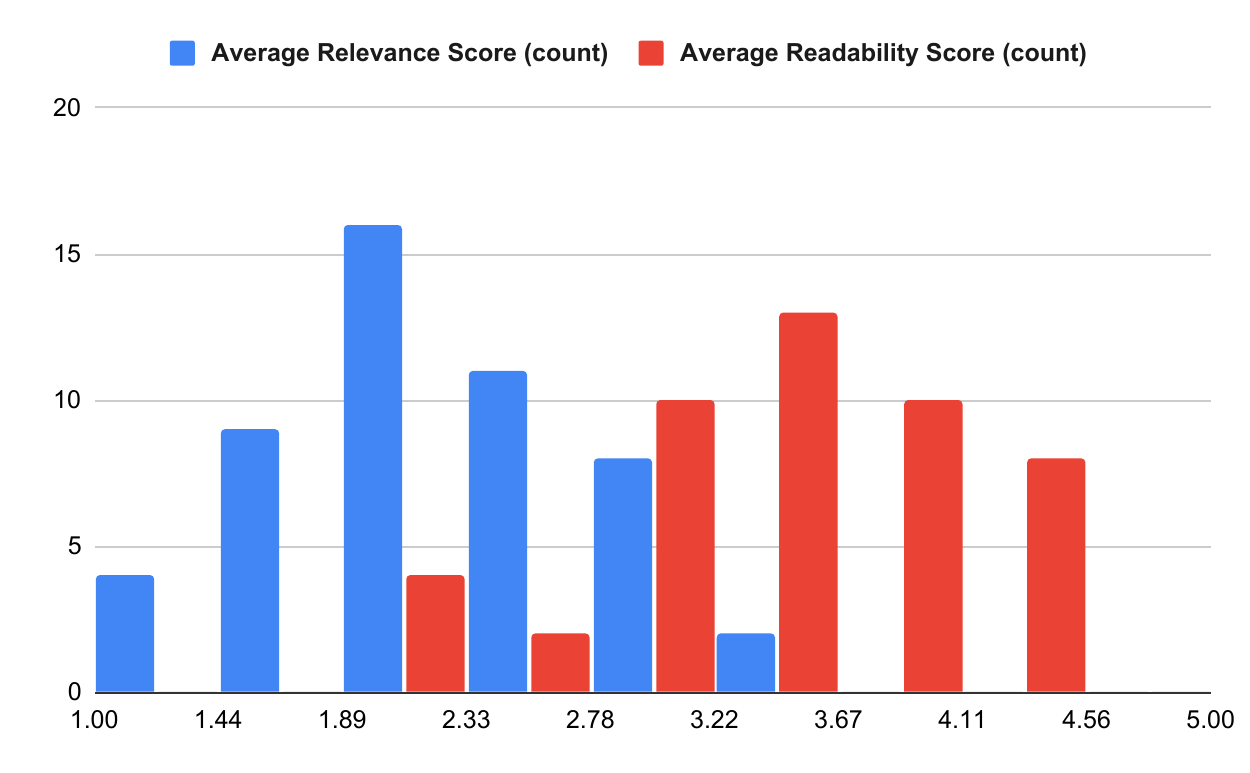}
    \caption{The distribution of average relevance and readability scores for summaries generated by GPT-4.} 
    \label{fig:score-gpt4}
\end{figure}

However, based on the results illustrated in Figure \ref{fig:scores-skt5} and Figure \ref{fig:score-gpt4}, we observe that although the summaries generated by SKT5SciSumm achieve better relevance scores, the readability scores require a significant improvement. In these figures, the average scores for each sample are calculated based on the scores given by the two human evaluators. Specifically, the average relevance score of the summaries generated by SKT5SciSumm is higher than 3, while the average relevance score of GPT-4 is lower than 3. This observation is further confirmed by the voting in task (i), where both evaluators favored the SKT5SciSumm text.

In Figure \ref{fig:scores-skt5}, we also note that the readability scores for SKT5SciSumm are quite evenly distributed between 2 and 4. In contrast, according to Figure \ref{fig:score-gpt4}, the GPT-4 summaries exhibit a narrower range of relevance scores, which range between 1 and 3. However, GPT-4 demonstrates the ability to generate summaries with better coherence and fluency, as indicated by higher readability scores. This suggests that GPT-4, as a world-class large language model, can produce well-written summaries even with minimal context.

\section{Example of our generated summaries among four models of T5}
\label{sec:appendix_a}
\begin{table*}[t]
\centering
\resizebox{\textwidth}{!}{%
\begin{tabular}{p{6.5in}}
\hline
\textbf{Reference}: Most recent approaches rely on convolutional neural networks (CNNs). cite , the work in cite is extended by adding a CNN to describe the posterior density of an object pose. A combination of using a CNN for object segmentation and geometry-based pose estimation is proposed in cite . PoseCNN cite uses a similar two-stage network, in which the first stage extracts feature maps from RGB input and the second stage uses the generated maps for object segmentation, 3D translation estimation and 3D rotation regression in quaternion format. Depth data and ICP are used for pose refinement. cite propose a three-stage, instance-aware approach for 6D object pose estimation. An instance segmentation network is first applied, followed by an encoder-decoder network which estimates the 3D object coordinates for each segment. The 6D pose is recovered with a geometric pose optimization step similar to cite. The approaches cite cite cite do not directly use CNN to predict the pose. Instead, they provide segmentation and other intermediate information, which are used to infer the object pose.                   \\ \hline
\textbf{T5-small}: In cite , the problem of \textbf{object pose} estimation is based on a \textbf{convolutional neural network (CNN)} that estimates the \textbf{3D translation} of an object by localizing its center in the image and predicting its distance from the camera. \textbf{PoseCNN} estimates the posterior density of a particular object pose with a Convolutional Neural Network (NCNN) which compares an observed and rendered image. This is done by describing a posterior density in a given image. cite proposed a new convolution neural network for \textbf{6D pose estimation}. In this paper, we propose a method that decomposes a 6D object pose. In a similar way, the YCB-Video dataset cite is used to estimate the pose of objects. The first step consists of two steps: cite and cite use a \textbf{segmentation network} to map objects to 3D surface positions, and thus eliminates surrounding clutter and occluders. This method does not require any post processing such as a single input image. The second step is a symmetric object pose estimate. In our work, we introduce a large scale video dataset for object pose prediction.                   \\ \hline
\textbf{T5-base}:
\textbf{PoseCNN} cite is a \textbf{convolutional neural network (CNN)} that estimates the \textbf{3D translation} of an object by localizing its center in the image and predicting its distance from the camera. This is done by describing the posterior density of a particular object pose with a CNN that compares an observed and rendered image. cite , a new Convolutional Neural Network for \textbf{6D object pose estimation} is introduced by cite and uses a recurrent neural network to estimate its posterior density. The first step localizes all known objects in an image using an \textbf{instance segmentation network}. The second step densely maps pixels to 3D object surface positions, so called object coordinates, using an \textbf{encoder-decoder network}, and hence eliminates object appearance. The third step decomposes the \textbf{6D pose using geometric optimization}. Then, the final step predicts the object pose using a novel loss function that enables the Pose-CNN to handle symmetric objects. The final step does not require any post-processing.                   \\ \hline
\textbf{T5-large}:
\textbf{PoseCNN} cite uses a \textbf{convolutional neural network (CNN)} to estimate the \textbf{3D translation} of an object by localizing its center in the image and predicting its distance from the camera. This is done by describing the \textbf{posterior density of a particular object pose with a CNN}  that compares an observed and rendered image. cite , a Convolutional Neural Network for \textbf{6D object pose estimation} is proposed. This method is based on the idea of localizing an object's center in an image and predicts its distance. The first step is to remove the surrounding clutter and occluders. The second and third step is used to predict the 6D pose. Then, the second step densely maps pixels to \textbf{3D object surface positions} using an \textbf{encoder-decoder network}, and hence eliminates object appearance. The third and final step is the final step, where the object coordinates are extracted from the image. The fourth and final pose is estimated using an iterative closest point algorithm. The final step of estimating the pose is performed using an \textbf{instance-aware pipeline} that decomposes the first two steps.                   \\ \hline
\textbf{T5-xl}:
\textbf{PoseCNN} in cite utilizes a \textbf{convolutional neural network (CNN)} to estimate the \textbf{3D translation} of an object by localizing its center in the image and predicting its distance from the camera. This involves describing the \textbf{posterior density of a particular object pose with a CNN}  that compares an observed and rendered image. Convolutional Neural Network in cite for \textbf{6D object pose estimation} is proposed. This method centers on the idea of localizing an object's center in an image and predicts its distance. The initial step involves removing the surrounding clutter and occluders. The subsequent two steps are employed to predict the 6D pose. In the second step, pixels are densely mapped to \textbf{3D object surface positions} using an \textbf{encoder-decoder network}, effectively eliminating object appearance. The third and final step, where object coordinates are extracted from the image, is the ultimate stage. The fourth and final pose is estimated using an iterative closest point algorithm. The pose estimation's concluding step is carried out using an \textbf{instance-aware} pipeline that decomposes the initial two steps.                   \\ \hline
\end{tabular}}
\caption{An example generated by fine-tuned T5 models on Multi-XScience corpus. The \textbf{highlighed} words are salient academic phrases.}
\label{tab:results_analysis}
\end{table*}

\end{document}